\documentclass[letterpaper, 10 pt, conference]{ieeeconf}  

\IEEEoverridecommandlockouts                              

\usepackage{graphicx} 
\usepackage{color}
\usepackage[ruled,linesnumbered]{algorithm2e}

\usepackage{amsmath}
\usepackage{amsfonts}
\usepackage{subcaption}
\usepackage{wrapfig}
\usepackage{url}
\expandafter\def\expandafter\normalsize\expandafter{%
    \normalsize
    \setlength\abovedisplayskip{2pt}
    \setlength\belowdisplayskip{5pt}
    \setlength\abovedisplayshortskip{2pt}
    \setlength\belowdisplayshortskip{5pt}
}

\DeclareMathOperator*{\argmin}{arg\,min}

\newcommand{\vx}{\mathbf{x}}
\newcommand{\vu}{\mathbf{u}}
\newcommand{\vU}{\mathbf{U}}
\newcommand{\loss}{\ell}
\newcommand{\trajinfo}{I}
\newcommand{\simloss}{\mathcal L}
\newcommand{\shaping}{\delta}
\newcommand{\data}{\mathcal D}
\newcommand{\pf}{\hat{f}}
\newcommand{\st}{\textrm{ s.t. }}
\newcommand{\hH}{\bar{H}}
\newcommand{\hu}{\bar{\vu}}
\newcommand{\algname}{HIMPC}
\newcommand{\empmu}{\hat{\mu}}
\newcommand{\empsig}{\hat{\Sigma}}
\newcommand{\priormu}{\mu_p}
\newcommand{\priorsig}{\Sigma_p}
\newcommand{\onlinediscount}{\beta}
\newcommand{\datapt}{\mathbf{p}}
\newcommand{\nth}{\textrm{th}}
\newcommand{\nn}{g}
\newcommand{\tind}{s}

\newcommand{\stt}{\vx_t}
\newcommand{\at}{\vu_t}
\newcommand{\Vxxt}{V_{\vx,\vx t}}
\newcommand{\Vxxtp}{V_{\vx,\vx t+1}}
\newcommand{\Vxtp}{V_{\vx t+1}}
\newcommand{\Vxt}{V_{\vx t}}
\newcommand{\Qyyt}{Q_{\vx\vu,\vx\vu t}}
\newcommand{\Qyt}{Q_{\vx\vu t}}
\newcommand{\costhesst}{\loss_{\vx\vu,\vx\vu t}}
\newcommand{\costgradt}{\loss_{\vx\vu t}}
\newcommand{\ddpdiscount}{}
\newcommand{\fyt}{f_{\vx\vu t}}
\newcommand{\Qxxt}{Q_{\vx,\vx t}}
\newcommand{\Quxt}{Q_{\vu,\vx t}}
\newcommand{\Quut}{Q_{\vu,\vu t}}
\newcommand{\Qut}{Q_{\vu t}}
\newcommand{\Qxt}{Q_{\vx t}}
\newcommand{\kpol}{k}
\newcommand{\Kpol}{K}

\renewcommand{\baselinestretch}{.97}

\title{\LARGE \bf
Learning from the Hindsight Plan -- Episodic MPC Improvement
}

\author{Aviv Tamar$^1$, Garrett Thomas$^1$, Tianhao Zhang$^1$, Sergey Levine$^1$, Pieter Abbeel$^{1,2,3}$
\thanks{$^1$EECS Department, UC Berkeley, 
$^2$ICSI, UC Berkeley, 
$^3$OpenAI}%
}

\begin{document}

\maketitle
\thispagestyle{empty}
\pagestyle{empty}

\begin{abstract}
Model predictive control (MPC) is a popular control method that has proved effective for robotics, among other fields.  MPC performs re-planning at every time step.  Re-planning is done with a limited horizon per computational and real-time constraints and often also for robustness to potential model errors. However, the limited horizon leads to suboptimal performance. In this work, we consider the iterative learning setting, where the same task can be repeated several times, and propose a policy improvement scheme for MPC. The main idea is that between executions we can, offline, run MPC with a longer horizon, resulting in a \emph{hindsight plan}.  To bring the next real-world execution closer to the hindsight plan, our approach learns to re-shape the original cost function with the goal of satisfying the following property: short horizon planning (as realistic during real executions) with respect to the shaped cost should result in mimicking the hindsight plan.  This effectively consolidates long-term reasoning into the short-horizon planning. We empirically evaluate our approach in contact-rich manipulation tasks both in simulated and real environments, such as peg insertion by a real PR2 robot.
\end{abstract}

\section{INTRODUCTION}
\label{sec:intro}
Model predictive control (MPC), also known as receding horizon control, is an effective model-based control method that is well-suited for complex dynamical systems, with a wide range of applications in robotics (see, e.g., \cite{camacho2013model,erez2013integrated}) among other domains.
In MPC, a model of system dynamics, either learned or known, is used to plan a locally optimal control policy for a \emph{limited horizon}, starting from the current state of the system. The first control in the plan is executed, and then re-planning is performed from the new state.

When the system dynamics are not known exactly, as is often the case in practice, MPC can be integrated with online system identification to simultaneously learn dynamics and plan controls.
This approach has been successfully applied in various robotic tasks, from aerial vehicle flight \cite{aswani2012extensions,chowdhary2013concurrent}, to contact-rich object manipulation \cite{lenz2015deepmpc,fu2016one}. In these applications, the limited horizon of MPC serves a dual purpose: maintaining tractability of the planning problem and mitigating error propagation during planning as a result of inaccurate models \cite{jiang2015dependence}. Indeed, MPC has been successfully applied to problems with challenging dynamics, such as cutting vegetables \cite{lenz2015deepmpc} and object manipulation \cite{fu2016one}.

One drawback of MPC is that planning with a limited horizon can lead to suboptimal policies.
For example, consider a task of navigating an environment with an obstacle. The MPC policy can only maneuver around the obstacle when it is within the planning horizon.
If the robot were to encounter the same task several times, we should naturally expect it to \emph{learn} to maneuver around the obstacle even before it enters the planning horizon. However, state-of-the-art MPC methods that only learn dynamics, such as \cite{lenz2015deepmpc} and \cite{fu2016one}, are prone to repeatedly produce suboptimal behavior in each episode of the task due to the limited horizon.

In this work, we propose a method that improves the MPC policies in episodic tasks. Our main insight is that, between episodes, we can revisit the MPC planning computations that were performed online, and recompute them \emph{offline} with a longer horizon, and with potentially better dynamics. 
The difference between this \emph{hindsight plan} and the actions that were actually performed can be used to drive a policy improvement \emph{between episodes}. In particular, we learn a neural network cost shaping for MPC using supervised learning, by minimizing this difference. The result is a method that incorporates long-term reasoning into MPC, while maintaining the benefits of short-horizon planning.

In comparison to previous policy improvement methods, which are typically based on value functions or policy gradients \cite{kober2013reinforcement}, our approach exploits the predictive nature of MPC to drive policy improvement, by contrasting the predicted actions with actions calculated in hindsight. This allows us to circumvent the difficulties of value function approximation and the sample inefficiency of policy gradients.

We show that our method effectively improves MPC policies for contact-rich manipulation tasks, such as peg insertion, in both simulated and real environments.

\begin{figure}[t!]
\centering
\includegraphics[width=0.45\textwidth,trim={5cm 6cm 2cm 3cm},clip]{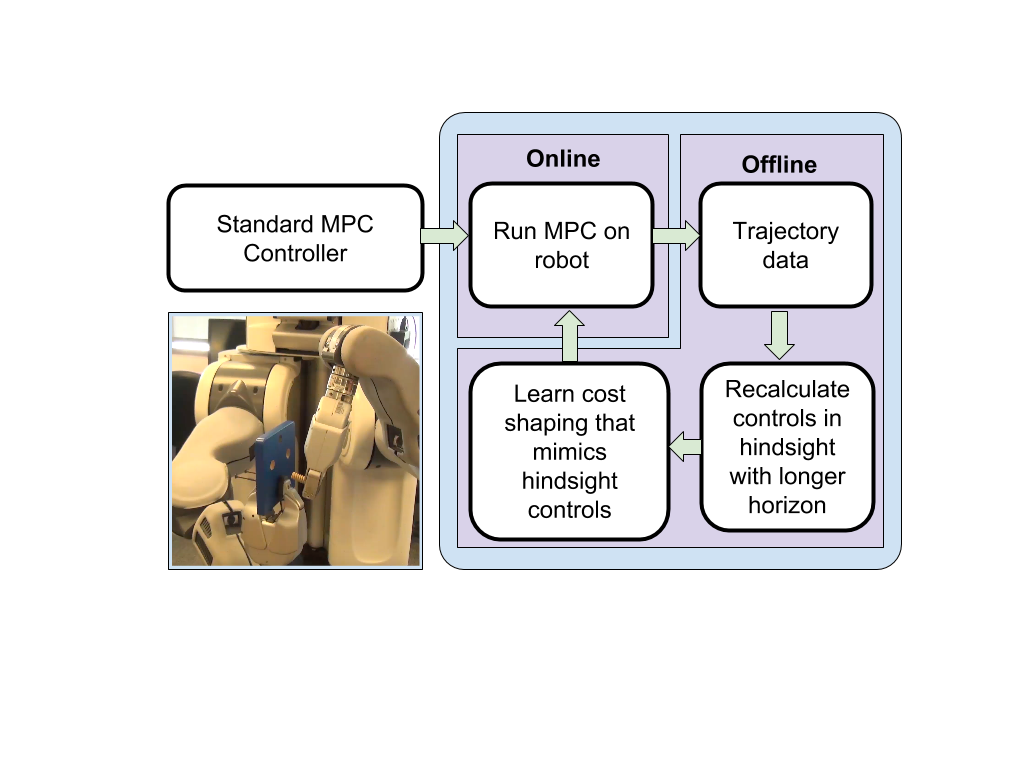}
\caption{
\small Right: A schematic of our method. After running an 
MPC controller with a limited horizon, we perform an offline calculation of the MPC on the same trajectory but with a longer horizon. This \emph{hindsight plan} is used to learn a cost shaping for the next execution of the MPC. Left: PR2 learning to insert a wooden peg into a hole. }
\label{fig:intro}
\vspace*{-1.8em}
\end{figure}
\section{RELATED WORK}
A standard approach for mitigating the limited planning horizon in MPC is to introduce a value function as the terminal cost, also known as \emph{infinite horizon MPC} \cite{chen1998quasi,erez2012infinite}. When the model is not fully known, or too large to calculate the value function, approximation techniques must be employed. However, it has been observed that standard value function approximation methods perform poorly when combined with MPC \cite{zhong2013value}. Zhong et el. \cite{zhong2013value} also proposed an improved approximation based on state discretization. However, that method does not scale to high-dimensional systems such as the 7 DoF robot arm considered in our experiments. We note that the cost shaping in our method can be set to have the form of a terminal value function. However, instead of learning directly from observed scalar costs, as typical for value function approximation \cite{kober2013reinforcement}, our method learns the cost shaping from the hindsight plan, which contains \emph{desired controls} and hence is much more informative. 

Our episodic formulation of improving the MPC controller is similar to the setting of iterative learning control (ILC)~\cite{moore2000iterative,longman2000iterative,bristow2006survey,wang2009survey}.
However, in ILC, the goal is typically to follow a reference trajectory, which is not suitable for the manipulation tasks we consider here. Our formulation allows the goal to be specified as a general optimal control problem. We note that a very recent work \cite{rosolia2016learning} proposed an extension of ILC that does not require a reference trajectory, and also uses a form of learning MPC, based on adding a value function to the MPC cost, which is calculated for all previously visited states. That approach, however, requires on a controller that can deterministically drive the system to any previously visited state, which does not hold for the real-world manipulation tasks we consider here.

Reinforcement planning \cite{zucker2012reinforcement} considers a discrete high-level planner for a continuous low-level control algorithm, and learns cost parameters of the planner using reinforcement learning methods. Our approach, in comparison, exploits the fact that our controller is based on MPC
to learn from the hindsight plan, and is not limited to discrete planning. We also note that our method is different from \emph{hindsight optimization} \cite{chong2000framework}, in which simulation is used to calculate a value function online.

Learning a cost function from observed controls is a form of inverse optimal control (IOC) \cite{rust1988maximum,abbeel2004apprenticeship}. While in IOC, the actions are assumed to be generated by an expert demonstrator, our method does not require such demonstrations, and the actions used for learning the cost shaping are generated by a planning algorithm which learns the model from interaction, as in model-based reinforcement learning (RL) \cite{deisenroth2011pilco,levine2014learning} and adaptive control \cite{aastrom2013adaptive}. Model-based RL is a standard approach in robotic control \cite{kober2013reinforcement}, achieving  state-of-the-art results in various domains, from helicopter flight \cite{abbeel2007application} to contact-rich manipulation
\cite{levine2015learning}. The use of MPC in RL allows for extremely data-efficient learning \cite{fu2016one}, which is especially important in robotics, where robot interaction time is often costly. In this work we improve the MPC controllers of \cite{fu2016one} for contact-rich manipulation tasks. However, our method is general, and can be applied to any adaptive MPC algorithm, and any domain for which MPC is suitable.


\section{PRELIMINARIES}
\label{sec:background}
In this work, we consider the standard episodic reinforcement learning or optimal control setting in discrete time with episode length $T$. We denote by $\vx_t$ the state at time $t$, and $\vu_t$ the control. The goal is to generate a control sequence $\vU_{0:T} = (\vu_0, \dots, \vu_T)$ that minimizes the total cost $\sum_{t=0}^T \loss_t (\vx_t, \vu_t)$ given an initial state $\vx_0$, where $\loss_t$ is a predefined task-specific scalar cost function, under dynamics given by $\vx_{t+1} = f(\vx_t, \vu_t)$. 
At time step $t$, adaptive MPC methods calculate the control $\vu_t$ by first estimating the dynamics model and then approximately solving for the optimal action. 

Concretely, let ${\data_t = \{(\vx_\tind, \vu_\tind, \vx_{\tind+1})\}_{s=0}^{t-1}}$ denote a dataset of the observed interaction with the system up to time $t$. Let $H \leq T$ denote the MPC horizon, which is a parameter of the algorithm. At time $t$, MPC\footnote{Note that this slightly non-standard notation of the MPC algorithm makes explicit the online dynamics learning.} first uses $\data_t$ to predict system dynamics $\pf_\tind^t$ for the next $H$ time steps, i.e. for ${t\,\leq\,\tind\,\leq\,t\!+\!H}$. Using the predicted dynamics, MPC then calculates the $H$-horizon optimal actions ${\vU_{t:t+H}^* = (\vu_t^*, \dots, \vu_{t+H}^*)}$, by solving%
\begin{small}
\begin{equation}
\label{eq:MPC_opt}
\begin{aligned}
& \argmin_{\vu_t,\dots,\vu_{t+H}} & & \sum_{\tind=t}^{t+H} \loss_\tind (\vx_\tind, \vu_\tind), \\
&\st & & \vx_{\tind+1} = \pf_\tind^t(\vx_\tind, \vu_\tind),\quad \forall \tind\!=\!t,\dots,t\!+\!H.
\end{aligned}
\end{equation} 
\end{small}%
The action $\vu_t = \vu_t^*$ is then taken, and the system transitions to a new state $\vx_{t+1}=f(\vx_t,\vu_t)$, from which the MPC optimization is repeated. When the episode ends, at $t=T$, the system is reset to the initial condition $\vx_0$ and the process iterates. To reduce clutter, we omit the episode index from the notation.

\section{THE HINDSIGHT PLAN}
\label{sec:hindsight}
In principle, if we know the true system dynamics $f$, and set $H$ equal to the episode length $T$, the solution of Eq.~\eqref{eq:MPC_opt} would lead to an optimal policy. However, in practice, both the computational burden of solving Eq.~\eqref{eq:MPC_opt}, and the error in dynamics prediction, necessitate the use of a shorter horizon $H < T$ (and often much shorter $H \ll T$) for online MPC optimization, which results in a \emph{suboptimal} policy. To mitigate this sub-optimality, we propose to revisit the planning computation \emph{after an episode has ended}, using a longer horizon, and potentially better dynamics, as we describe in this section. This additional planning computation, which we term the \emph{hindsight plan}, will allow us to improve our MPC policy, as we shall later describe.

Let $\hH$ denote the horizon of the hindsight plan, where typically\footnote{With perfect dynamics, setting $\hH=T$ would be optimal. However, with an inaccurate dynamics model, setting $\hH<T$ may be preferred.} $\hH \geq H$. The \emph{hindsight action} $\hu_t$ at time $t$ is defined as $\hu_t = \hu_t^*$, where $\bar{\vU}_{t:t+\hH}^*=(\hu_t^*,\dots,\hu_{t+\hH}^*)$ is the solution to the following \emph{hindsight planning problem}:%
\begin{equation}\label{eq:HMPC_opt}
\begin{aligned}
&\argmin_{\vu_t,\dots,\vu_{t+\hH}} 
& & \sum_{\tind=t}^{t+\hH} \loss_\tind (\vx_\tind, \vu_\tind), \\
& \st & & \vx_{\tind+1} = \pf_\tind^\tind(\vx_\tind, \vu_\tind),\quad \forall s=t,\dots, t\!+\!\hH,
\end{aligned}
\end{equation}
where the initial state $\vx_t$ is the state observed at time $t$ during the original MPC execution, and the terms $\pf_\tind^\tind$ are the same terms that were already calculated for the original MPC planning problem \eqref{eq:MPC_opt}. Henceforth, we collectively term $\vx_t$ and $\pf_\tind^\tind$ as the MPC trajectory information $\trajinfo\doteq\{\vx_t,\pf_\tind^\tind\}$.

Let us emphasize the differences between the hindsight planning problem~\eqref{eq:HMPC_opt} and the original MPC planning problem \eqref{eq:MPC_opt}. First, we plan with a longer horizon $\hH$. Second, at each time step $t$, we use the dynamics predictions $\pf_t^t,\dots,\pf_{t+\hH}^{t+\hH}$ that were calculated at times $t,\dots,t+\hH$ of the online MPC execution, as opposed to the predictions that were calculated at time $t$, $\pf_t^t,\dots,\pf^{t}_{t+\hH}$  in Eq. \eqref{eq:MPC_opt}. These predictions were not available at time $t$ during the online MPC execution, as they use observations \emph{from later time steps} in the episode\footnote{Potentially, we can use the data set at the end of the episode $\data_T$ to re-estimate the dynamics at every time step, and obtain even better dynamics predictions. However, we found it sufficient to use the dynamics that were already calculated during the online MPC execution.}.

The main assumption underlying our approach is that the hindsight plan produces improved actions compared to the original MPC execution. This assumption is motivated by the improved  dynamics prediction in \eqref{eq:HMPC_opt}, which is based on future data and, more importantly, the much longer planning horizon.
Thus, in the hindsight plan, the two main sources of MPC sub-optimality are removed. We also note that calculating the hindsight plan typically requires more computation than the original MPC planning during the episode execution. However, the real-time requirement of MPC can be relaxed, because it can be computed offline, and the calculation can be performed concurrently and in parallel for all $t$.

\section{MPC POLICY IMPROVEMENT}
In this section, we show how the hindsight plan can be used to drive a policy improvement algorithm, by using it for learning a \emph{cost shaping} for MPC. 
The idea is to learn a cost shaping that encourages the online MPC solution to be \emph{more similar to the solution in hindsight}. 

Let us revisit the MPC optimization problem, and add to the original cost a cost shaping term, $\shaping_\tind(\vx_\tind, \vu_\tind, \theta)$, parametrized by some vector $\theta$,%
\begin{equation}\label{eq:Shaped_MPC_opt}
\begin{split}
 \argmin_{\vu_t,\dots,\vu_{t\!+\!H}} & \!\sum_{\tind=t}^{t+H}\! \loss_\tind (\vx_\tind, \vu_\tind \!) \!+\! \shaping_\tind(\vx_\tind, \vu_\tind, \theta), \\
    \st & \vx_{\tind+1} = \pf_\tind^t(\vx_\tind, \vu_\tind),\quad \forall \tind\!=\!t,\dots,t\!+\!H
\end{split}
\raisetag{12pt}
\end{equation} 
where, similarly to Eq.~\eqref{eq:HMPC_opt}, the trajectory information $\trajinfo = \{\vx_t,\pf_\tind^t\}$ contains the terms that were calculated for the original MPC planning problem \eqref{eq:MPC_opt}.
The cost-shaping term $\shaping_\tind(\vx_\tind, \vu_\tind, \theta)$ can be represented, for example, by a neural network parametrized by $\theta$. We denote by $\vu_t(\theta)$ the first action in the solution of \eqref{eq:Shaped_MPC_opt} for time $t$. That is, $\vu_t(\theta)$ denotes the action MPC \emph{would have taken} during the episode, had the cost-shaping parameter been $\theta$. We also denote by $\vu_t^0$ the first action in the solution of \eqref{eq:Shaped_MPC_opt} with $\shaping_\tind(\vx_\tind, \vu_\tind, \theta)\equiv0$, that is, the solution with no cost shaping applied. 

We aim to learn a parameter $\theta$ that encourages the MPC solution of Eq. \eqref{eq:Shaped_MPC_opt} to be similar to the the hindsight plan \eqref{eq:HMPC_opt}. We therefore propose to learn $\theta$ by minimizing the similarity loss $\simloss$, defined as%
\begin{small}
\begin{equation}\label{eq:cost_learning_opt}
\simloss(\theta) = \sum_{t=0}^T \|\vu_t(\theta) - \hu_t\|^2 + \lambda \|\vu_t(\theta) - \vu_t^0\|^2,
\end{equation}
\end{small}%
where $\lambda \geq 0$ acts as a regularization term that controls the change from the online MPC policy. Note that the dependence of $\vu_t(\theta)$ on $\theta$ can be quite complex, as it encapsulates the solution of the planning problem \eqref{eq:Shaped_MPC_opt}. A similar problem is encountered in IOC \cite{rust1988maximum,abbeel2004apprenticeship}, and indeed, our approach can be seen as performing IOC with the hindsight plan replacing the expert demonstration. For planning algorithms that can be represented as a computation graph \cite{schulman2015gradient}, such as linear quadratic regulator (LQR) and value iteration \cite{Tamar16}, the loss $\simloss$ can be minimized efficiently using gradient based methods, such as L-BFGS \cite{liu1989limited}. We also note that our method can be used to learn additional parameters of the planning algorithm, such as the dynamics and discount factor, by simply adding them to the parameter vector $\theta$. In this work, however, we focus on learning cost shaping. 

Furthermore, if $\shaping_\tind(\vx_\tind, \vu_\tind, \theta)$ is set to $0$ for all $\tind < t + H$, the cost shaping becomes a terminal value function, which is exactly the infinite horizon MPC formulation \cite{chen1998quasi,erez2012infinite}. A crucial difference in our approach, however, is that we learn the parameters $\theta$ by minimizing the similarity loss $\simloss$, which directly measures the difference between the online and hindsight actions. Standard value function methods first approximate a value function, and then approximately solve the planning problem with that value function, resulting in two sources of approximation error, possibly explaining their poor performance observed in previous work \cite{zhong2013value}.

The solution to Eq.~\eqref{eq:cost_learning_opt} provides us with the parameters of a shaping cost that causes the MPC controller to mimic the hindsight controller, and results in a single step of policy improvement.  However, once we run the MPC with the shaped cost in the system, we obtain a new trajectory, which we can perform hindsight planning on. This results in an iterative policy improvement algorithm, which we term \emph{hindsight iterative MPC} (\algname). Let $i\geq 0$ denote an iteration of the algorithm, and let $\theta_i$ denote the cost-shaping parameters used for the MPC execution at iteration $i$, where we assume that the first iteration $i=0$ is executed without any cost shaping. We denote by $\simloss_{i}(\theta)$ the similarity loss optimization problem \eqref{eq:cost_learning_opt}, where the trajectory information $\trajinfo_i$ is calculated from execution of the MPC controller at iteration $i$, with cost-shaping parameter $\theta_{i}$. We calculate $\theta_{i+1}$ by minimizing $\sum_{k=0}^i \simloss_{k}(\theta)$, which assures that we aggregate the hindsight information obtained at previous iterations. 

The general \algname\ algorithm is summarized in Algorithm \ref{alg:HIMPC}. So far, we have not discussed the specific algorithms used for dynamics predictions and planning in MPC \eqref{eq:MPC_opt}, nor the functional form of the cost shaping, and in principle, \algname\ can be combined with any adaptive MPC method such as \cite{lenz2015deepmpc} and \cite{fu2016one}. However, the specific choice of algorithm and cost shaping will determine the computational complexity of solving the cost-shaped MPC planning \eqref{eq:Shaped_MPC_opt} in real time. In addition, the tractability of minimizing the similarity loss \eqref{eq:cost_learning_opt} depends on the specific algorithm and cost shaping structure. In the next section, we describe a particular implementation of \algname\ based on LQR planning and Gaussian Mixture Model (GMM) dynamics. We also propose a neural-network (NN) cost shaping form that is both efficient to optimize and interpretable.

\begin{algorithm}
\DontPrintSemicolon
Run original MPC controller \eqref{eq:MPC_opt}, and collect trajectory information $\trajinfo_0$ \;

Calculate hindsight plan \eqref{eq:HMPC_opt} with $\trajinfo_0$ \;

Find cost-shaping that mimics hindsight controller  $\theta_0 = \argmin_\theta \simloss_0(\theta)$ \;

\For{i=1,2,\dots}{
Run MPC controller with cost-shaping parameter $\theta_{i-1}$, and collect trajectory information $\trajinfo_i$ \;

Calculate hindsight plan \eqref{eq:HMPC_opt} with  $\trajinfo_i$ \;

Solve $\theta_i = \argmin_\theta \sum_{k=0}^i \simloss_{k}(\theta)$ \;
}
\caption{HIMPC Algorithm}
\label{alg:HIMPC}
\end{algorithm}

\section{AN LQR IMPLEMENTATION OF HIMPC}
\label{sec:lqr}
In this section we describe a particular implementation of the \algname\ algorithm, based on LQR planning and GMM dynamics learning. We are inspired by the work of \cite{fu2016one}, in which similar methods were used within MPC for effectively performing contact-rich manipulation tasks. In addition, we propose a novel cost shaping structure, that is both efficient to optimize, and has an intuitive interpretation as learning `way points'. 

The online dynamics adaptation algorithm of \cite{fu2016one} is based on a Bayesian approach for estimation, where the recent observations within an episode are used to estimate a linear dynamics model, using observations from previous episodes as a Bayesian prior. This produces, for every time step $t$, a time-varying linear dynamics model of the form  $\vx_{\tind+1} = \pf_\tind^t(\vx_\tind, \vu_\tind) = A_\tind^t \vx_{\tind} + B_\tind^t \vu_{\tind}$, for $\tind = t,\dots,t+H$. The method is further described in Appendix \ref{sec:appendix_dynamics}.

As the loss function $\loss_\tind$, we use a quadratic loss of the form $\loss_\tind(\vx_\tind,\vu_\tind) = (\vx_\tind-\vx^*)^\top Q_\tind (\vx_\tind-\vx^*) + \vu_\tind^\top R_\tind \vu_\tind$, where $\vx^*$ is some goal state of the system, and $Q_\tind$ and $R_\tind$ are a positive semi-definite and positive definite matrices, respectively\footnote{Extending our approach to non-quadratic loss functions is straightforward, by using a second-order Taylor expansion. See, e.g., \cite{erez2013integrated} for details.}. Such a cost function is standard for many control tasks \cite{anderson2007optimal}, and is particularly  suitable for the manipulation experiments we consider, where the task is specified as moving the robot end-effector to some goal position, such as pushing a peg into a hole. 

With linear dynamics and a quadratic loss, the MPC planning problem \eqref{eq:MPC_opt} becomes a standard LQR problem \cite{anderson2007optimal}, for which a solution can be calculated efficiently by dynamic programming, as described Appendix \ref{sec:appendix_LQR}. The mapping from the trajectory information $\{\vx_t,\pf_\tind^t\}$ to the action $\vu_t$ in this case can be written as the sequence of matrix multiplications and inversions in the LQR solution. This mapping can be represented as a computation graph, and the gradient $\partial \vu_t / \partial \vx^*$ can be easily calculated using modern automatic differentiation packages such as Theano \cite{theano2016}.

We are now ready to introduce our cost shaping formulation. At time $t$, for $\tind = t,\dots,t+H$, we consider a cost shaping of the form:
\begin{small}%
\begin{equation}\label{eq:cost_structure}
    \shaping_\tind(\vx_\tind, \vu_\tind, \theta) = \nn(\vx_t, \theta)^\top Q_\tind \vx_\tind,
\end{equation}
\end{small}%
where $\nn(\vx_t, \theta)$ is a neural network that has the current state $\vx_t$ as its input, $\theta$ as its weights, and a vector-valued output with the same dimensionality as $\vx_t$. Note that adding a linear term to a quadratic form is equivalent to changing the center of the quadratic, up to a constant. Therefore, the shaped cost can be written as: %
\begin{small}
\begin{equation}\label{eq:cost_structure_quadratic}
\begin{split}
    &\loss_\tind(\vx_\tind,\vu_\tind) + \shaping_\tind(\vx_\tind, \vu_\tind, \theta)= \\ &\left(\vx_\tind-\hat{\vx}^*(\vx_t, \theta)\right)^\top Q_\tind \left(\vx_\tind-\hat{\vx}^*(\vx_t, \theta)\right) 
    + \vu_\tind^\top R_\tind \vu_\tind \!+\! \text{const},    
\end{split}
\raisetag{28pt}
\end{equation}
\end{small}%
where $\hat{\vx}^*(\vx_t, \theta) = \vx^* + \nn(\vx_t, \theta)$ is a \emph{modified goal position}. Thus, our cost shaping has the intuitive interpretation of modifying the goal position, which can be thought of as learning state-dependent way points. Note that $\nn$ depends on the current state observation $\vx_t$, and not on $\vx_\tind$, therefore solving LQR with the cost in \eqref{eq:cost_structure_quadratic} is equivalent to solving the original LQR, just with a different $\vx^*$, making the solution similarly tractable. 

In addition, calculating the gradient $\partial \vu_t / \partial \theta = \partial \vu_t / \partial \hat{\vx}^* \cdot \partial \hat{\vx}^* / \partial \theta$ is also tractable. This gradient can be used for minimizing the similarity loss \eqref{eq:cost_learning_opt} with standard optimization algorithms such as L-BFGS \cite{liu1989limited}. Note that the gradient $\partial \nn / \partial \vx$ is not required since $\nn$ depends on $\vx_t$, which is part of the trajectory information $\trajinfo$.

\subsection*{An Illustrative Example}\label{ssec:illustrative_exmple}
In this section we discuss an application of \algname\ to a simple 2D navigation task with obstacles. The goal of this example is to illustrate the function of the hindsight plan, and the learned cost shaping. Further quantitative analysis of this experiment is presented in Section \ref{sec:experiments}.

\begin{figure}[t]
      \centering
      \includegraphics[trim={0 0 0 0},clip, width=0.4\textwidth]{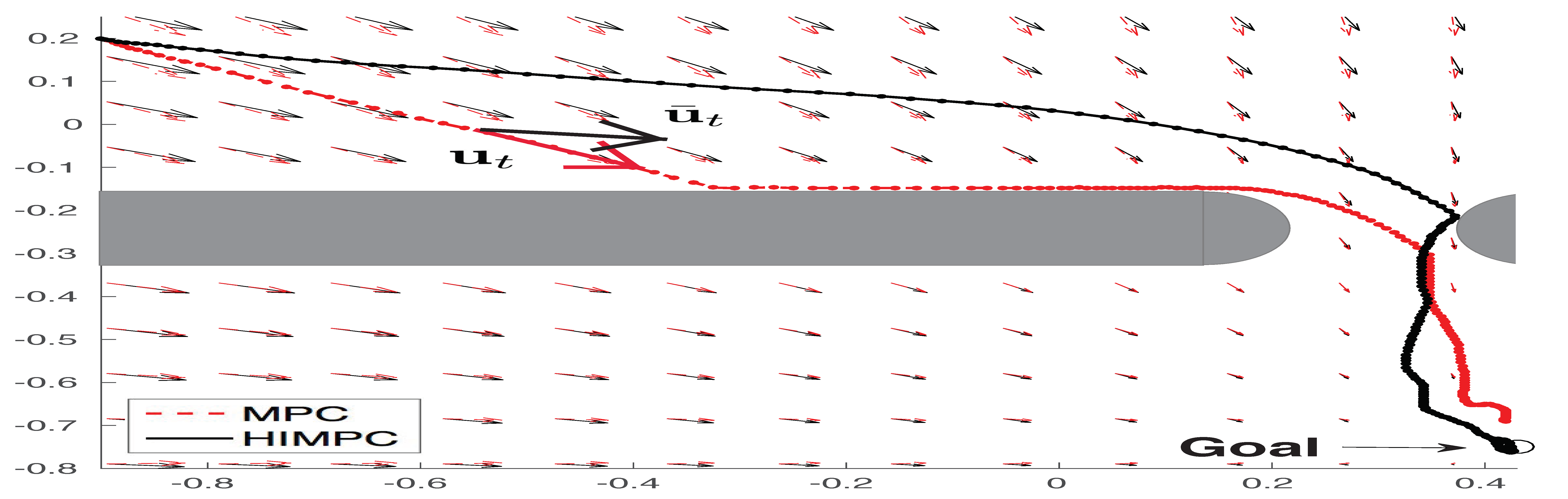}
      \caption{
      \small 
      2D navigation between obstacles. The task is to navigate a particle in 2D between the gray obstacles to the goal position. Dashed-red line: MPC controller. Solid-black line: \algname\ controller. Wide arrows show the action MPC chose at a particular time $\vu_t$, and the corresponding hindsight action $\hu_t$.
      The quiver plot shows the direction to the (shaped) goal position. Note that for \algname, the modified goal position for states at the top left is shifted to the right, leading to a trajectory that correctly navigates into the opening between the obstacles.}
      \label{fig:2D_obstacles}
      \vspace{-1.5em}
      \end{figure}

Consider the 2D navigation task depicted in Figure \ref{fig:2D_obstacles}. A particle with mass needs to navigate to a goal position by using vertical and horizontal forces. The particle may collide with the impenetrable gray colored obstacles, which when in contact, apply normal and frictional forces to the particle. 

Ideally, when starting from the initial position shown in Figure \ref{fig:2D_obstacles}, the particle should navigate to the opening between the obstacles and from there continue to the goal. Such a plan would minimize the total cost in this domain. However, when a MPC policy is applied, the controller first navigates to the obstacle location which is closest to the goal, as shown in the dashed-red line in Figure \ref{fig:2D_obstacles}, since its limited horizon planning (here $H=10$) and imperfect dynamics model fail to take into account the future obstacle collision. Only after experiencing the obstacle, the controller navigates alongside it towards the goal. While the initial MPC policy is able to solve the task, its solution is clearly sub-optimal. 

The wide red and black arrows in Figure \ref{fig:2D_obstacles} show the action that MPC selected, $\vu_t$, and the action that the hindsight plan (with $\hH=30$ in this case) has chosen, $\hu_t$, at a particular time step. Due to the better dynamics prediction and longer horizon, the hindsight plan correctly predicts the future collision, and takes an appropriate action.

The solid-black line in Figure \ref{fig:2D_obstacles} shows the trajectory performed by the \algname\ algorithm, after 5 episodes of learning. The controller learned to first navigate to the opening, as desired. In addition, we visualize the cost shaping by plotting a black quiver plot of the direction to the modified goal position $\hat{\vx}^*(\vx, \theta) - \vx$. Note how the modified goal position at the beginning of the trajectory orients towards the opening, and note the difference with the red quiver plot, which shows the direction to the original goal $\vx^* - \vx$.

\section{PRACTICAL IMPROVEMENTS OF HIMPC}
\label{sec:technical_improvements}

As with most machine learning algorithms, and neural networks in particular, a successful application of the method requires some technical know-how \cite{lecun2012efficient,glorot2010understanding}. In this section we report several technical procedures that we found to improve the performance of \algname\ in our experiments.

{\em a) Add control noise to the MPC:}
This helps the MPC controller get around `local minima'
in the trajectory by random exploration, and the \algname\ then learns a cost shaping that consolidates this trajectory improvement. In particular, we used the exploration scheme of \cite{fu2016one} in our experiments.

{\em b) Collect several trajectories in each iteration:}
We found that collecting several roll-outs of the same shaped MPC controller at each iteration (lines 1 and 5 in Algorithm \ref{alg:HIMPC}) stabilizes the neural network training. 

{\em c) Wait for successful MPC runs before initiating cost-shaping:}
When the standard MPC fails in the task due to a bad initial dynamics model, its trajectories do not contain enough useful knowledge for the hindsight planning. We therefore wait until several successful trajectories occur before starting \algname. We found measuring success by a threshold on the final distance to the target to perform well. While this condition is task dependent, in our experiments we did not find it to be sensitive to the threshold magnitude.  

{\em d) Early stopping of cost-shaping}
The learned cost shaping can potentially direct the controller to a different goal position than $\vx^*$. This often happens in early iterations, when not enough successful trajectories have been observed, and can cause the algorithm to destabilize, as it no longer receives successful trajectories. A solution to this problem is to turn off the cost shaping (during the run) if the shaped cost has converged, while the original cost has not. Such a fix guarantees that \algname\ is as stable as the original MPC.

\section{EXPERIMENTS}
\label{sec:experiments}
\begin{figure*}[t]
\vspace*{1em}
\captionsetup[subfigure]{aboveskip=0em,belowskip=0em}
\centering
\begin{subfigure}[t]{0.33\textwidth}
\includegraphics[width=\textwidth,trim=1em 0 5em 4em,clip]{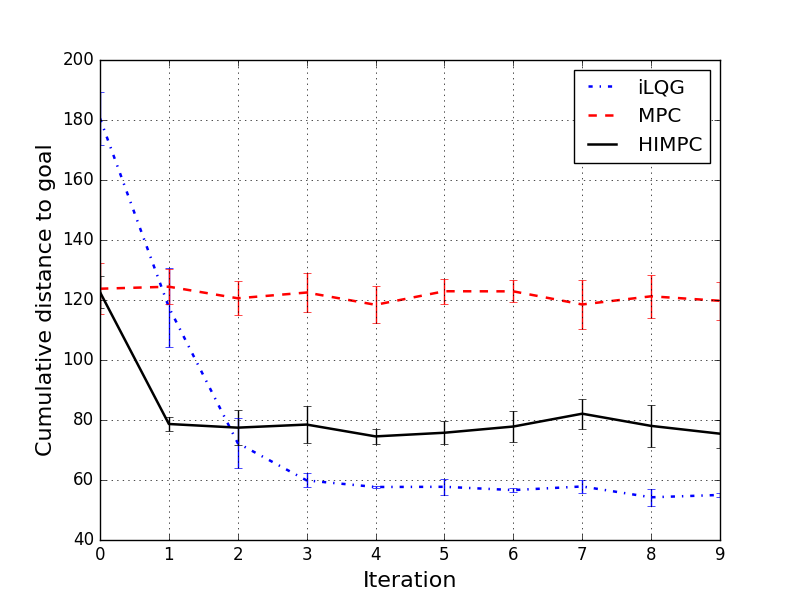}
\caption{2D simulation}
\label{fig:mjc-2d-insertion}
\end{subfigure}~
\begin{subfigure}[t]{0.33\textwidth}
\includegraphics[width=\textwidth,trim=1em 0 5em 4em,clip]{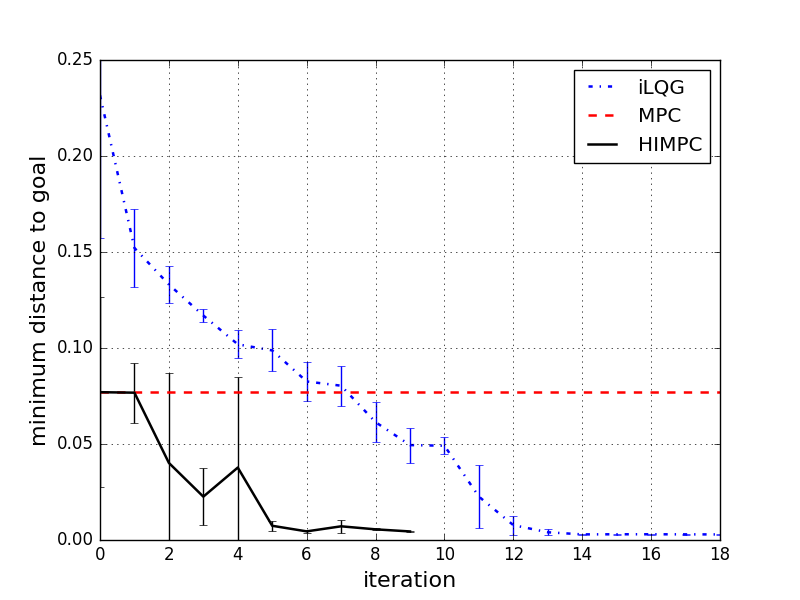}
\caption{3D simulation}
\label{fig:mjc-3d-insertion}
\end{subfigure}~
\begin{subfigure}[t]{0.33\textwidth}
\includegraphics[width=\textwidth,trim=1em 0 5em 4em,clip]{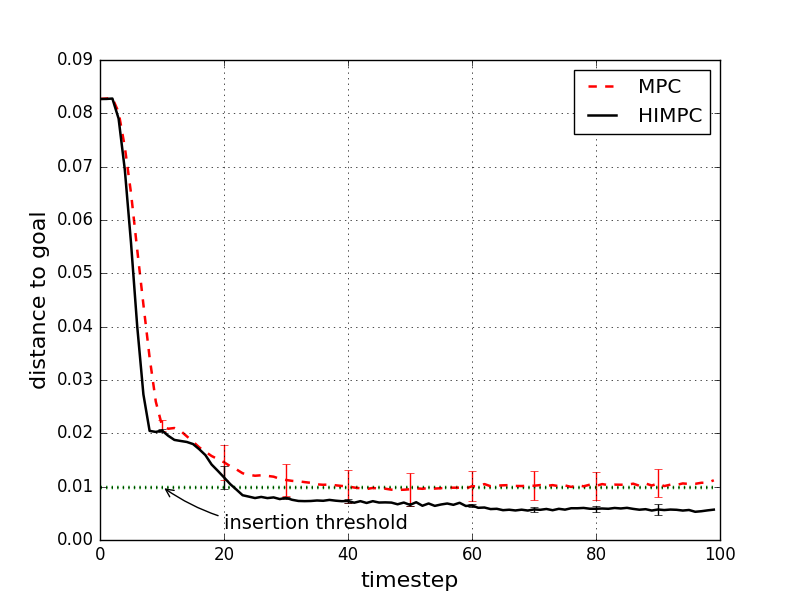}
\caption{real PR2}
\label{fig:pr2-insertion}
\end{subfigure}
\caption{
\small
Experimental results. (a): Performance plots for the 2D navigation task. (b): Performance plots for the simulated 3D peg insertion task. We show the minimal distance from the target, averaged over the trajectories in the episode. \algname\ significantly improves the baseline MPC performance, and converges faster than iLQG. (c): Performance plots for the peg insertion task with the real PR2. \algname\ inserts the peg faster, and with better precision.
}
\vspace{-1.5em}
\end{figure*}

\begin{figure}[t]
\captionsetup[subfigure]{aboveskip=0em,belowskip=0em}
\centering
\begin{subfigure}[t]{0.34\textwidth}
\includegraphics[width=\textwidth]{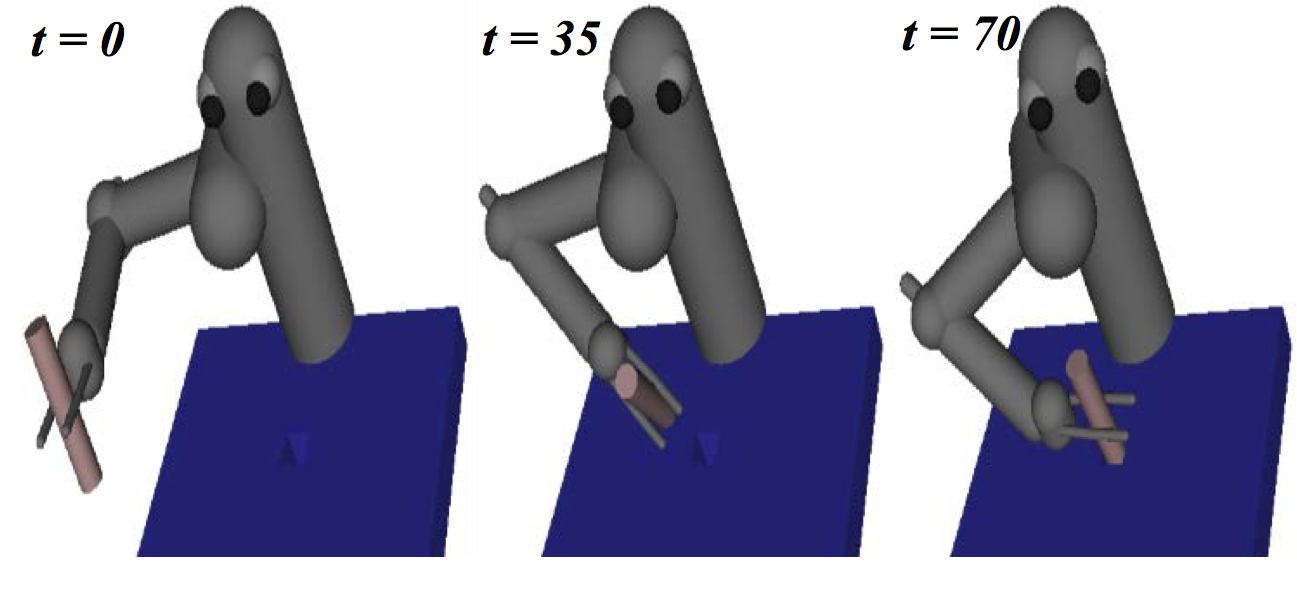}
\caption{\algname, simulated cylindrical peg insertion}
\label{fig:mjc-3d-insertion-himpc}
\end{subfigure}
\begin{subfigure}[t]{0.45\textwidth}
\includegraphics[width=\textwidth]{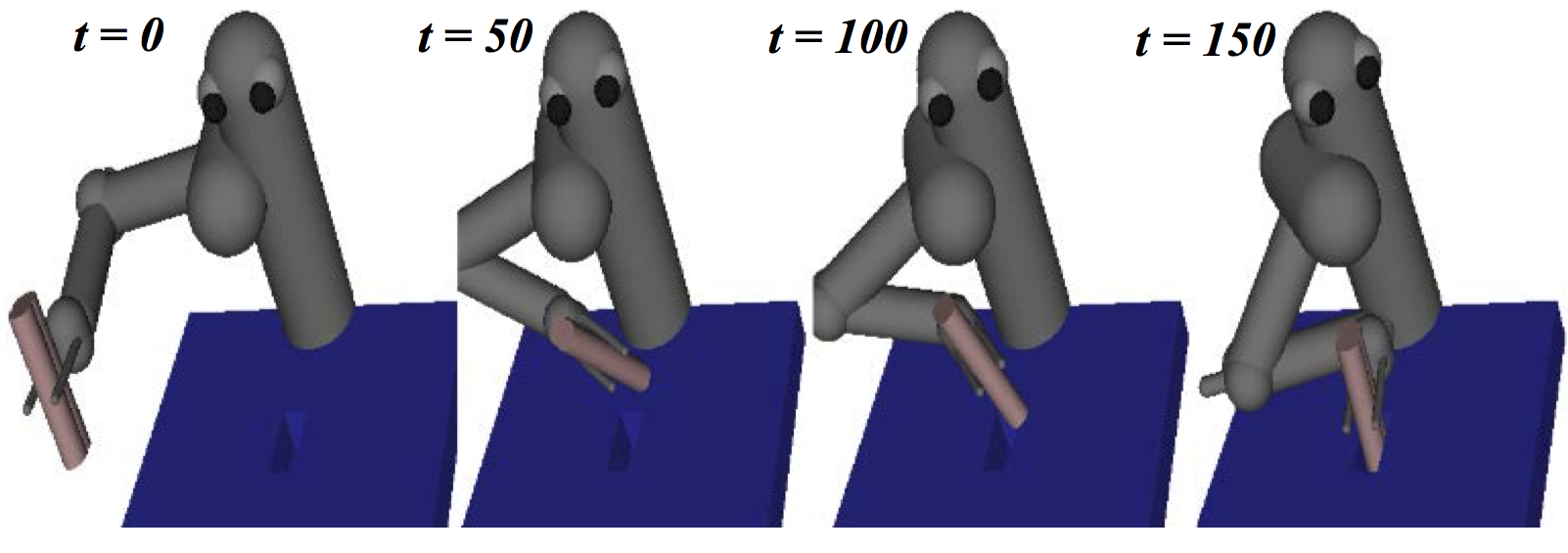}
\caption{\algname, simulated oblong peg insertion}
\label{fig:mjc-3d-insertion-mpc}
\end{subfigure}
\begin{subfigure}[t]{0.45\textwidth}
\includegraphics[width=\textwidth]{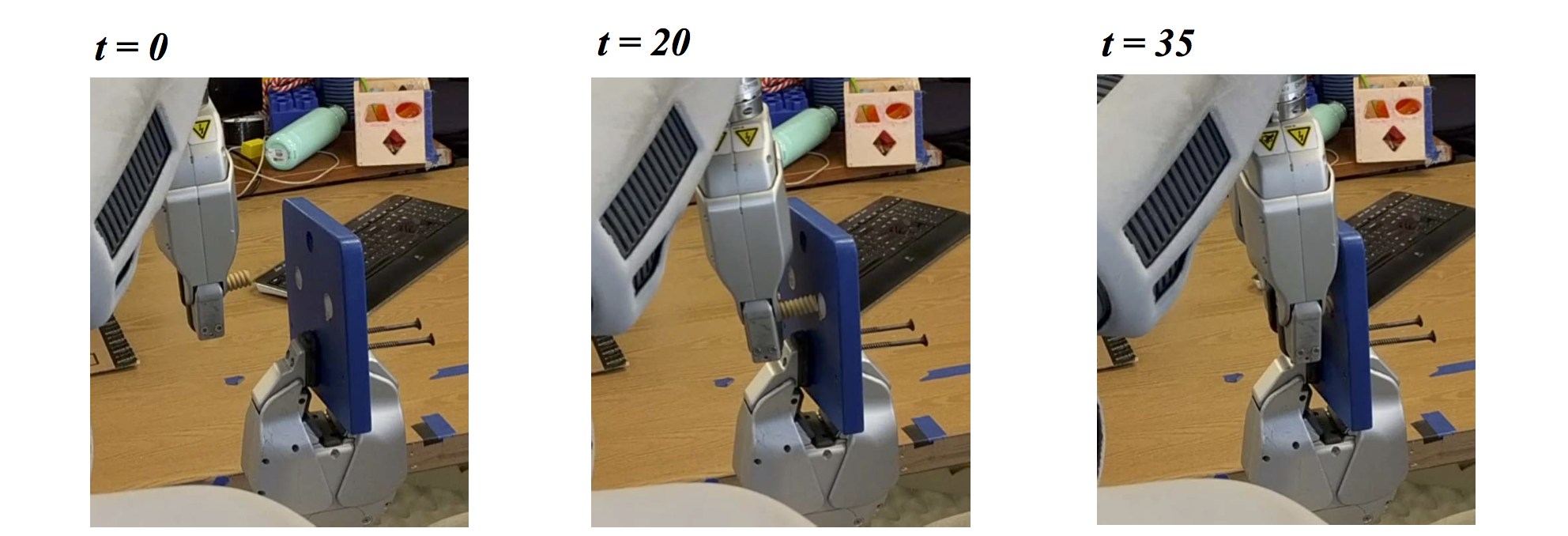}
\caption{HIMPC, real PR2 peg insertion}
\label{fig:pr2-3d-insertion-himpc}
\end{subfigure}
\begin{subfigure}[t]{0.45\textwidth}
\includegraphics[width=\textwidth]{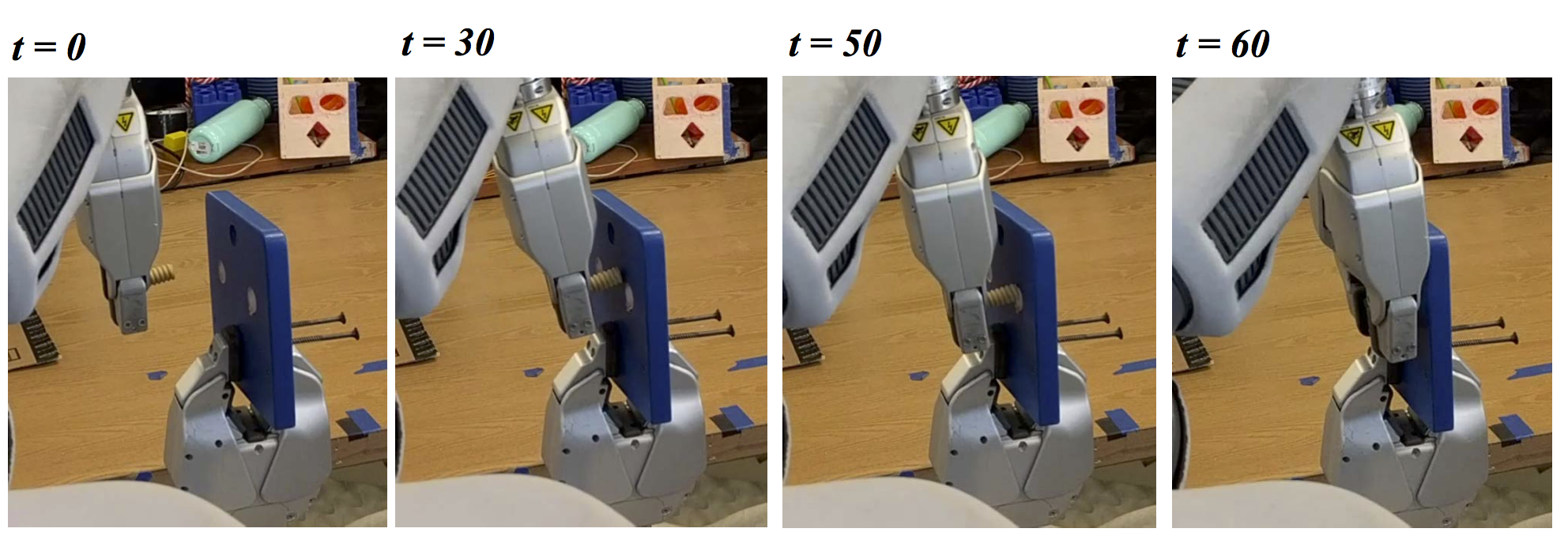}
\caption{MPC, real PR2 peg insertion}
\label{fig:pr2-insertion-mpc}
\end{subfigure}
\caption{
\small
Visualization of \algname\ and MPC in peg insertion tasks. (a,b): The \algname\ policy for the simulated peg insertion tasks. Note the different arm orientation for the cylindrical (a) and oblong (b) peg. (c,d) The \algname\ policy and baseline MPC policy for the real peg insertion task. Note that MPC first reaches an intermediate point on the plate, while \algname\ directly reaches the hole.
}
\end{figure}

\begin{figure}[t]
\centering
\includegraphics[width=0.45\textwidth]{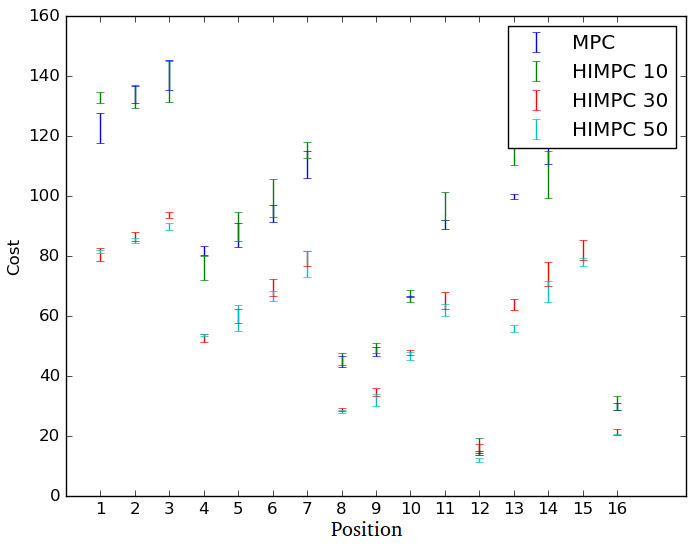}
\caption{
\small
Generalization and the effect of $\bar{H}$. Performance (cumulative distance to goal, averaged over 10 trials with standard deviation error bars) of MPC and \algname\ on the 2D navigation task, starting from 16 different initial positions, for various settings of $\bar{H}$. Note that performance consistently improves with a longer horizon for all test positions. These test positions were not used during training, and demonstrate the generalization capability of the neural network.
}
\label{fig:generalization}
\end{figure}

In this section, we experimentally evaluate the \algname\ algorithm in simulated and real robot experiments. In this work, we focus on contact-rich manipulation, following the work of Fu et al.~\cite{fu2016one}, though our method can be applied to other tasks where MPC is applicable. In particular, we focus on various insertion tasks, which are important for assembly, and for which the improvement of \algname\ over standard MPC can be easily visualized. 


In our evaluation, we aim to answer the following two questions:
\begin{enumerate}
    \item Can \algname\ improve upon standard MPC?
    \item Can \algname\ improve upon a standard episodic model-based RL approach (i.e., without MPC)?
\end{enumerate}

For our simulations, in addition to the 2D navigation task discussed in Section \ref{ssec:illustrative_exmple}, we consider two variants of a peg insertion task. All our simulations were performed using the MuJoCo physics engine \cite{todorov2012mujoco},
and our code, which is based on the guided policy search repository \cite{fzftm-gpsi-16}, will be made available. Additionally we experiment on a peg insertion task with a real PR2 robot. In both simulated and real experiments, we apply direct torque control at $20$Hz. For the simulated experiments, the state space is 26-dimensional, consisting of the positions and angular velocities of the 7 joints, and positions and velocities of 2 points on the end-effector. For the real robot experiment, an additional end-effector point was added, resulting in 32 state dimensions.

In our evaluation, we compare \algname\ to the MPC method of \cite{fu2016one}, and, as a baseline, to iLQG -- a state-of-the-art model-based RL method \cite{levine2014learning}.
Our dynamics model uses a GMM prior, combined with online dynamics adaptation \cite{fu2016one}, as further described in Appendix \ref{sec:appendix_dynamics}. The iLQG method of  \cite{levine2014learning} uses a similar GMM for a dynamics prior, but combines with time-varying linear dynamics, and therefore constitutes a fair comparison.
The same quadratic cost function was used for all algorithms. We note that the optimization in iLQG is performed over the full episode, and should in principal converge to a (locally) optimal solution. However, MPC is expected to be much more sample-efficient, as was demonstrated in \cite{fu2016one}.

In terms of computation time, hindsight planning  demands were comparable to system reset times between episodes, and had negligible effect. However, solving the optimization problem in \eqref{eq:cost_learning_opt} was much slower, requiring several minutes of computation, since our Theano-based implementation~\cite{theano2016} cannot differentiate a matrix inverse for multiple samples in parallel. We expect future automatic differentiation packages to substantially reduce the computation time.

\subsection{Simulated 2D Navigation}\label{ssec:particle}
This task, as depicted in Figure~\ref{fig:2D_obstacles}, requires moving a particle through an opening towards a goal position in 2D space. The available controls apply horizontal and vertical forces to the particle, and obstacles effect friction and normal forces. Episodes are $200$ time-steps long, and three trajectories are executed at each iteration. The horizon $H$ and hindsight horizon $\hH$ were chosen as $10$ and $30$, respectively. The cost-shaping neural network $\nn$ consists of two fully connected hidden layers, each with 25 units and $\tanh$ activations.
The inputs to $\nn$ are the particle position and velocity.

The dynamics GMM prior was initialized from a single episode with a random initial policy.
In addition, after each episode we update the prior with the samples from the most recent episode, to potentially improve the baseline MPC algorithm between iterations.

Our performance evaluation is the cumulative distance to the goal over the entire episode, which measures the speed of reaching the goal. In Figure~\ref{fig:mjc-2d-insertion}, we plot the averaged performance over four runs using different random seeds. As may be observed, \algname\ significantly improves over standard MPC. Interestingly, updating the dynamics prior did not significantly improve the original MPC (as can be seen by relatively similar performance across episodes). This is since in this simple domain, the dynamics prior of the first episode is already good enough. Note that the baseline iLQG reaches a better solution than MPC. This is expected, as iLQG solves the problem with the full horizon and therefore requires more samples than MPC.

\textbf{Generalization and the effect of $\bar{H}$.}
The \algname\ algorithm learns to improve performance when iteratively starting the task from the same initial position. An important consideration, however, is its sensitivity to the change of initial position.
In this experiment we demonstrate that the neural network cost shaping can generalize to initial positions that were not trained on.

Using the simulated 2D domain described in section \ref{ssec:particle}, we ran \algname\ using samples collected from five different initial positions. We then evaluated performance when starting from 16 different test positions, arranged in a $4 \times 4$ grid. We evaluated \algname\ for various values of $\bar{H}$, as well as standard MPC, on the test positions, and present the results (averaged across 10 test trials) in Figure~\ref{fig:generalization}.
Observe that as $\bar{H}$ increases, the performance of \algname\ improves consistently for all test positions. Furthermore, when $\bar{H} > H$, \algname\ outperforms standard MPC, even when run from initial positions that were not seen during training. This result shows that the benefit of planning with a longer horizon is indeed captured by the learned cost shaping, and also demonstrates the generalization capability of the neural network.


\subsection{Simulated 3D Peg Insertion}
\label{ssec:peg_insertion}
In this task, a 7-Dof robot arm
must insert a cylindrical peg into a tight-fitting rectangular hole, as depicted in Figure~\ref{fig:mjc-3d-insertion-himpc} \& \ref{fig:mjc-3d-insertion-mpc}. The controls are the torques applied to the 7 joints, and the state space consists of the positions and angular velocities of these joints and the 3D positions and velocities of 2 points on the end-effector, which are hereafter referred to as the \emph{EE points}.
The goal is specified by the coordinates of the EE points (but not the joints), when the peg is fully inserted in the hole. Thus, there is a rotational degree of freedom in the goal position, since there are only 2 EE points. This non-trivial task requires solving both a kinematic problem, and a control problem with complex contact dynamics, and has been used as a benchmark in previous studies \cite{zhang2016learning}.

Episodes are $400$ time-steps long, and $3$ trajectories are executed at each iteration. The horizon $H$ is $10$, while the hindsight horizon $\hH$ was chosen as $60$. The NN $\nn$ had 2 fully connected hidden layers of sizes $[100,25]$ with $\tanh$ activations, and its inputs were the positions of the EE points. 
The dynamics GMM prior was initialized from a single episode with a random policy. In this experiment we did not update the prior after the first episode, as we observed it to decrease performance. We attribute this to the GMM method, which is sensitive to the choice of clusters and the input distribution, and can degrade performance when additional samples are added. Using dynamics models such as neural networks \cite{fu2016one} or Gaussian processes \cite{pan2016adaptive} could potentially improve the dynamics learning, as we plan to investigate in future work. We emphasize, however, that improving MPC dynamics should also improve the performance of \algname.

To make a fair comparison with iLQG, which learns time-dependent linear dynamics models, we executed $6$ trajectories of $200$ time-steps at each iLQG iteration. This results in a better performance of iLQG, while the number of samples (total time steps) at each iteration is the same as \algname.

Our performance evaluation is the minimal distance to the goal over the episode, which measures the success of inserting the peg. In Figure~\ref{fig:mjc-3d-insertion}, we plot our results, averaged over the trajectories in the episode. As may be observed, \algname\ significantly improves over the standard MPC, which is equivalent to the performance of \algname\ in the first episode. In this domain, \algname\ solves the task with significantly fewer samples than iLQG. 

\subsection{Simulated 3D Oblong Peg Insertion}\label{ssec:oblong_peg}
In this experiment, we demonstrate how \algname\ with random exploration noise in the control,
can learn additional structure in the task, and represent it in the shaped cost.

This task is similar to the previous peg insertion task, with a difference that the peg and hole have an oblong shape, requiring a particular orientation of the peg for a successful insertion. As before, the goal is specified by the coordinates of the 2 EE points, when the peg is fully inserted in the hole. Thus, the cost function \emph{does not contain information} about the correct orientation for solving the task. With sufficient exploration noise, standard MPC can sometimes solve this task. Our goal is to show that \algname\ can consolidate the information from these `lucky' runs, and learn a cost shaping that guides the peg into the correct orientation. 

As before, episodes are $400$ time-steps long, and $3$ trajectories are executed at each iteration. The horizon $H$ is $10$, while the hindsight horizon $\hH$ was chosen as $60$, and the NN $\nn$ had 2 fully connected hidden layers of sizes $[100,25]$ with $\tanh$ activations. In this case, however, we added to the NN shaping-cost an additional EE point, which was not used for the original cost specification. Thus, the cost-shaping has the capacity to represent an orientation of the peg. We emphasize that this additional EE point \emph{was not} part of the MPC cost, and the orientation can only be learned from the hindsight plan on `lucky' trajectories.

In Table \ref{table:oblong} we report the success rate of peg insertion for MPC, and \algname\ after 6 episodes of learning. We also report on success or failure when no control noise is added. After 6 episodes, \algname\ has learned a cost shaping that orients the peg correctly, and succeeds in the task, even without control noise. MPC on the other hand, cannot solve the task without exploration noise. This behavior can be further visualized in the supplemental video\footnote{\url{https://sites.google.com/site/himpchindsightplan/}}.

\begin{table}[h]
\vspace*{-1em}
\captionsetup{aboveskip=-0.2em,belowskip=-0.2em}
\caption{Success Rates for Oblong Peg Insertion}
\label{table:oblong}
\begin{center}
\begin{tabular}{|c|c|c|}
\hline
& W/ Noise & W/O Noise\\
\hline
MPC & 4/10 & Fail \\
\hline 
\algname\ & 8/10 & Success \\
\hline
\end{tabular}
\end{center}
\vspace*{-2em}
\end{table}

\subsection{Real PR2 Experiments}
We evaluated our method on a peg insertion task with the PR2 robot. The robot is tasked with inserting a small wooden peg into a hole in a wooden plate. The task specification is similar to the simulated peg insertion of Section \ref{ssec:peg_insertion}, and the goal position is specified by the position of 3 EE points, when the peg is fully inserted in the hole. 

We constructed a GMM dynamics prior from $40,000$ samples of actions in free space, collected by running the iLQG algorithm for $30$ minutes, reaching random goal positions. Such a large data set was required for a stable MPC control. In addition, we used a shorter hindsight horizon than in simulation $\hH\!=\!30$, to account for the less accurate dynamics.

In Figure~\ref{fig:pr2-insertion}, we plot the distance of the EE points from their goal position, for the original MPC controller, and for \algname\ after $3$ episodes of learning, averaged over $5$ executions of the controller. Similarly to the 2D task described above, the MPC controller in this task tries to approach the goal position in a straight line, and bumps into the wooden plate. It then glides into the opening to insert the peg. \algname, on the other hand, learns to directly insert the peg into the opening, and therefore reaches the goal faster. This behavior can be visualized in the supplemental video$^5$.

\section{CONCLUSION}
\label{sec:conclusion}
In this work we introduced a new approach for policy improvement in repeated tasks. Rather than using value functions or policy gradients -- the traditional drivers of policy improvement in RL -- our method employs an online MPC policy, and suggests improved actions based on an offline hindsight calculation once an episode has ended.

We demonstrated a significant improvement over standard MPC on several complex manipulation tasks with contacts, and a notable improvement in sample efficiency over state-of-the-art model based RL.

In future work we intend to investigate the use of different dynamics prediction models in our method, and applications in different robotics domains such as quadrotors. 
In addition, the explicit use of the prediction error as a driver for policy improvement could potentially be used in different RL algorithms.

\appendices
\section{}
\label{sec:appendix_dynamics}
\vspace{-0.5em}
\textbf{Dynamics Prediction:} We used the dynamics prediction method of \cite{fu2016one}. An exponential moving average of the observations is maintained throughout the episode $\empmu_t \leftarrow \onlinediscount \empmu_{t-1} + (1 - \onlinediscount) \datapt_t$, where $\datapt_t = [\vx_{t-1} ; \vu_{t-1} ; \vx_t]$ is the $t^\nth$ observation and $\onlinediscount$ is a discounting factor that causes the model to forget old data. A similar exponential moving covariance is maintained by $\empsig_t \leftarrow \onlinediscount \empsig_{t-1} + (1 - \onlinediscount) \datapt_t \datapt_t^\top$. These within-episode dynamics are mixed with a prior model of dynamics $(\priormu, \priorsig)$, by $\mu = \alpha_1 \empmu + (1-\alpha_1) \priormu$, and $\Sigma = \alpha_2 \priorsig + \alpha_3 \empsig + \alpha_4 (\priormu-\empmu)(\priormu-\empmu)^\top$, where the prior is a GMM fit to samples from previous episodes, and the mixing coefficients $\alpha_1,\alpha_2,\alpha_3,\alpha_4$ are described in \cite{fu2016one}. A linear dynamics model $\vx_{t+1} = A_t \vx_{t} + B_t \vu_{t}$ is obtained by assuming a multivariate Gaussian distribution for $[\vx_{t} ; \vu_{t} ; \vx_{t+1}]$ with mean $\mu$ and covariance $\Sigma$, and conditioning $\vx_{t+1}$ on $[\vx_{t},\vu_{t}]$, producing a Gaussian distribution with mean $[A_t, B_t]$. 
We refer to \cite{fu2016one} for the full details and theoretical motivation of this algorithm. To predict the dynamics for time $\tind = t+1$, the previous MPC policy (for time $t-1$) is used to predict the current action $\hat{\vu}_t$, and the predicted next state is $\hat{\vx}_{t+1} = A_t \vx_{t} + B_t \hat{\vu}_{t}$. The prior is queried for the dynamics of state $\hat{\vx}_{t+1}$, and mixed with the current dynamics estimate as described above, to produce a linear dynamics model $\vx_{\tind+1} = A_\tind^t \vx_{\tind} + B_\tind^t \vu_{\tind}$. This process can be repeated iteratively for $\tind = t+2,\dots,t+H$, producing a time-varying linear dynamics model for the horizon $H$.
\section{}
\label{sec:appendix_LQR}
\textbf{LQR: }Consider a linear dynamics system $\vx_{t+1} = f_t(\vx_t, \vu_t) \doteq A_t \vx_{t} + B_t \vu_t$, and a quadratic loss function $\loss_t(\vx_t,\vu_t) = (\vx_t-\vx^*)^\top D_t (\vx_t-\vx^*) + \vu_t^\top R_t \vu_t$. The Q-function and value function are both quadratic, given by%
\begin{footnotesize}
\begin{align*}
V(\stt) &= \frac{1}{2}\stt^\top\Vxxt\stt + \stt^\top\Vxt + \text{const} \\
Q(\stt,\at)\! &= \!\frac{1}{2}[\stt;\! \at]^\top \! \Qyyt[\stt;\! \at] \!+\! [\stt;\! \at]^\top \! \Qyt \!+\! \text{const},
\end{align*}
\end{footnotesize}%
Using dynamics programming \cite{anderson2007optimal}:%
\begin{footnotesize}
\begin{align*}
\Qyyt &= \costhesst + \ddpdiscount\fyt^\top \Vxxtp\fyt \\
\Qyt &= \costgradt + \ddpdiscount\fyt^\top \Vxtp \\
\Vxxt &= \Qxxt - \Quxt^\top \Quut^{-1}\Quxt \\
\Vxt &= \Qxt - \Quxt^\top \Quut^{-1}\Qut.
\end{align*}
\end{footnotesize}%
The optimal control law is linear, and given by $\vu_t = \kpol_t + \Kpol_t \vx_t$, where $\Kpol_t = -\Quut^{-1} \Quxt$ and $\kpol_t = -\Quut^{-1} \Qut$. 
\section*{ACKNOWLEDGMENT}
This work was supported in part by Siemens, the DARPA SIMPLEX program, an NSF CAREER Award, and an ONR Young Investigator Award. Aviv Tamar was partially funded by the Viterbi Scholarship, Technion. Tianhao Zhang was supported by a UC Berkeley EECS Departmental Fellowship. The authors thank Ramu Chandra, Karthik Kappaganthu, and Juan L. Aparicio for fruitful discussions, and Justin Fu for useful advice, and for sharing his adaptive MPC code.


\begin{thebibliography}{10}
\providecommand{\url}[1]{#1}
\csname url@rmstyle\endcsname
\providecommand{\newblock}{\relax}
\providecommand{\bibinfo}[2]{#2}
\providecommand\BIBentrySTDinterwordspacing{\spaceskip=0pt\relax}
\providecommand\BIBentryALTinterwordstretchfactor{4}
\providecommand\BIBentryALTinterwordspacing{\spaceskip=\fontdimen2\font plus
\BIBentryALTinterwordstretchfactor\fontdimen3\font minus
  \fontdimen4\font\relax}
\providecommand\BIBforeignlanguage[2]{{%
\expandafter\ifx\csname l@#1\endcsname\relax
\typeout{** WARNING: IEEEtran.bst: No hyphenation pattern has been}%
\typeout{** loaded for the language `#1'. Using the pattern for}%
\typeout{** the default language instead.}%
\else
\language=\csname l@#1\endcsname
\fi
#2}}

\bibitem{camacho2013model}
E.~F. Camacho and C.~B. Alba, \emph{Model predictive control}.\hskip 1em plus
  0.5em minus 0.4em\relax Springer Science \& Business Media, 2013.

\bibitem{erez2013integrated}
T.~Erez, K.~Lowrey, Y.~Tassa, V.~Kumar, S.~Kolev, and E.~Todorov, ``An
  integrated system for real-time model predictive control of humanoid
  robots,'' in \emph{Humanoids}, 2013.

\bibitem{aswani2012extensions}
A.~Aswani, P.~Bouffard, and C.~Tomlin, ``Extensions of learning-based model
  predictive control for real-time application to a quadrotor helicopter,'' in
  \emph{ACC}, 2012, pp. 4661--4666.

\bibitem{chowdhary2013concurrent}
G.~Chowdhary, M.~M{\"u}hlegg, J.~P. How, and F.~Holzapfel, ``Concurrent
  learning adaptive model predictive control,'' in \emph{Advances in Aerospace
  Guidance, Navigation and Control}.\hskip 1em plus 0.5em minus 0.4em\relax
  Springer, 2013, pp. 29--47.

\bibitem{lenz2015deepmpc}
I.~Lenz, R.~Knepper, and A.~Saxena, ``Deepmpc: Learning deep latent features
  for model predictive control,'' in \emph{Robotics Science and Systems}, 2015.

\bibitem{fu2016one}
J.~Fu, S.~Levine, and P.~Abbeel, ``One-shot learning of manipulation skills
  with online dynamics adaptation and neural network priors,'' \emph{IROS},
  2016.

\bibitem{jiang2015dependence}
N.~Jiang, A.~Kulesza, S.~Singh, and R.~Lewis, ``The dependence of effective
  planning horizon on model accuracy,'' in \emph{AAMAS}.\hskip 1em plus 0.5em
  minus 0.4em\relax International Foundation for Autonomous Agents and
  Multiagent Systems, 2015, pp. 1181--1189.

\bibitem{kober2013reinforcement}
J.~Kober, J.~A. Bagnell, and J.~Peters, ``Reinforcement learning in robotics: A
  survey,'' \emph{International Journal of Robotics Research}, 2013.

\bibitem{chen1998quasi}
H.~Chen and F.~ALLGoWER, ``A quasi-infinite horizon nonlinear model predictive
  control scheme with guaranteed stability,'' \emph{Automatica}, vol.~34,
  no.~10, pp. 1205--1217, 1998.

\bibitem{erez2012infinite}
T.~Erez, Y.~Tassa, and E.~Todorov, ``Infinite-horizon model predictive control
  for periodic tasks with contacts,'' \emph{Robotics: Science and systems},
  p.~73, 2012.

\bibitem{zhong2013value}
M.~Zhong, M.~Johnson, Y.~Tassa, T.~Erez, and E.~Todorov, ``Value function
  approximation and model predictive control,'' in \emph{2013 IEEE Symposium on
  Adaptive Dynamic Programming and Reinforcement Learning (ADPRL)}.\hskip 1em
  plus 0.5em minus 0.4em\relax IEEE, 2013, pp. 100--107.

\bibitem{moore2000iterative}
K.~L. Moore and J.-X. Xu, \emph{Iterative learning control}.\hskip 1em plus
  0.5em minus 0.4em\relax Taylor \& Francis, 2000.

\bibitem{longman2000iterative}
R.~W. Longman, ``Iterative learning control and repetitive control for
  engineering practice,'' \emph{International journal of control}, vol.~73,
  no.~10, pp. 930--954, 2000.

\bibitem{bristow2006survey}
D.~A. Bristow, M.~Tharayil, and A.~G. Alleyne, ``A survey of iterative learning
  control,'' \emph{IEEE Control Systems}, 2006.

\bibitem{wang2009survey}
Y.~Wang, F.~Gao, and F.~J. Doyle, ``Survey on iterative learning control,
  repetitive control, and run-to-run control,'' \emph{Journal of Process
  Control}, vol.~19, no.~10, pp. 1589--1600, 2009.

\bibitem{rosolia2016learning}
U.~Rosolia and F.~Borrelli, ``Learning model predictive control for iterative
  tasks,'' \emph{arXiv preprint arXiv:1609.01387}, 2016.

\bibitem{zucker2012reinforcement}
M.~Zucker and J.~A. Bagnell, ``Reinforcement planning: {RL} for optimal
  planners,'' in \emph{ICRA}.\hskip 1em plus 0.5em minus 0.4em\relax IEEE,
  2012, pp. 1850--1855.

\bibitem{chong2000framework}
E.~K. Chong, R.~L. Givan, and H.~S. Chang, ``A framework for simulation-based
  network control via hindsight optimization,'' in \emph{CDC}, 2000.

\bibitem{rust1988maximum}
J.~Rust, ``Maximum likelihood estimation of discrete control processes,''
  \emph{SIAM Journal on Control and Optimization}, vol.~26, no.~5, pp.
  1006--1024, 1988.

\bibitem{abbeel2004apprenticeship}
P.~Abbeel and A.~Y. Ng, ``Apprenticeship learning via inverse reinforcement
  learning,'' in \emph{ICML}.\hskip 1em plus 0.5em minus 0.4em\relax ACM, 2004,
  p.~1.

\bibitem{deisenroth2011pilco}
M.~Deisenroth and C.~E. Rasmussen, ``{PILCO}: A model-based and data-efficient
  approach to policy search,'' in \emph{ICML}, 2011, pp. 465--472.

\bibitem{levine2014learning}
S.~Levine and P.~Abbeel, ``Learning neural network policies with guided policy
  search under unknown dynamics,'' in \emph{NIPS}, 2014.

\bibitem{aastrom2013adaptive}
K.~J. {\AA}str{\"o}m and B.~Wittenmark, \emph{Adaptive control}.\hskip 1em plus
  0.5em minus 0.4em\relax Courier Corporation, 2013.

\bibitem{abbeel2007application}
P.~Abbeel, A.~Coates, M.~Quigley, and A.~Y. Ng, ``An application of
  reinforcement learning to aerobatic helicopter flight,'' \emph{NIPS}, 2007.

\bibitem{levine2015learning}
S.~Levine, N.~Wagener, and P.~Abbeel, ``Learning contact-rich manipulation
  skills with guided policy search,'' in \emph{ICRA}, 2015.

\bibitem{schulman2015gradient}
J.~Schulman, N.~Heess, T.~Weber, and P.~Abbeel, ``Gradient estimation using
  stochastic computation graphs,'' in \emph{NIPS}, 2015, pp. 3528--3536.

\bibitem{Tamar16}
A.~Tamar, Y.~Wu, G.~Thomas, S.~Levine, and P.~Abbeel, ``Value iteration
  networks,'' \emph{CoRR}, vol. abs/1602.02867, 2016.

\bibitem{liu1989limited}
D.~C. Liu and J.~Nocedal, ``On the limited memory {BFGS} method for large scale
  optimization,'' \emph{Mathematical programming}, 1989.

\bibitem{anderson2007optimal}
B.~D. Anderson and J.~B. Moore, \emph{Optimal control: linear quadratic
  methods}.\hskip 1em plus 0.5em minus 0.4em\relax Courier Corporation, 2007.

\bibitem{theano2016}
{Theano Development Team}, ``{Theano: A {Python} framework for fast computation
  of mathematical expressions},'' \emph{arXiv e-prints}, vol. abs/1605.02688.

\bibitem{lecun2012efficient}
Y.~A. LeCun, L.~Bottou, G.~B. Orr, and K.-R. M{\"u}ller, ``Efficient
  backprop,'' in \emph{Neural networks: Tricks of the trade}.\hskip 1em plus
  0.5em minus 0.4em\relax Springer, 2012.

\bibitem{glorot2010understanding}
X.~Glorot and Y.~Bengio, ``Understanding the difficulty of training deep
  feedforward neural networks.'' in \emph{Aistats}, vol.~9, 2010, pp. 249--256.

\bibitem{todorov2012mujoco}
E.~Todorov, T.~Erez, and Y.~Tassa, ``Mujoco: A physics engine for model-based
  control,'' in \emph{2012 IEEE/RSJ International Conference on Intelligent
  Robots and Systems}.\hskip 1em plus 0.5em minus 0.4em\relax IEEE, 2012, pp.
  5026--5033.

\bibitem{fzftm-gpsi-16}
\BIBentryALTinterwordspacing
C.~Finn, M.~Zhang, J.~Fu, X.~Tan, Z.~McCarthy, E.~Scharff, and S.~Levine,
  ``Guided policy search code implementation,'' 2016. [Online]. Available:
  \url{http://rll.berkeley.edu/gps}
\BIBentrySTDinterwordspacing

\bibitem{zhang2016learning}
M.~Zhang, Z.~McCarthy, C.~Finn, S.~Levine, and P.~Abbeel, ``Learning deep
  neural network policies with continuous memory states,'' in
  \emph{ICRA}.\hskip 1em plus 0.5em minus 0.4em\relax IEEE, 2016, pp. 520--527.

\bibitem{pan2016adaptive}
Y.~Pan, X.~Yan, E.~Theodorou, and B.~Boots, ``Adaptive probabilistic trajectory
  optimization via efficient approximate inference,'' \emph{arXiv preprint
  arXiv:1608.06235}, 2016.

\end{thebibliography}

\end{document}